\pdfoutput=1
\documentclass[11pt]{article}

\usepackage[final]{acl}
\usepackage{times}
\usepackage{latexsym}

\usepackage[T1]{fontenc}
\usepackage[utf8]{inputenc}

\usepackage{microtype}

\usepackage{inconsolata}

\usepackage{graphicx}

\usepackage{enumitem}

\title{TokenSmith: Streamlining Data Editing, Search, and Inspection for Large-Scale Language Model Training and Interpretability}

\usepackage{listings}
\usepackage{xcolor}
\usepackage{svg}
\usepackage{graphicx}     % For \includegraphics
\usepackage{subcaption}   % For subfigures using subfigure environment

\definecolor{codegray}{rgb}{0.5,0.5,0.5}
\definecolor{codeblue}{rgb}{0.2,0.3,0.9}
\definecolor{codeorange}{rgb}{0.8,0.4,0}
\definecolor{backcolor}{rgb}{0.97,0.97,0.97}

\author{
 \textbf{Mohammad Aflah Khan\textsuperscript{1}\thanks{Equal contribution}},
 \textbf{Ameya Godbole\textsuperscript{2}\footnotemark[1]},
 \\
 \textbf{Johnny Tian-Zheng Wei\textsuperscript{2}},
 \textbf{Ryan Wang\textsuperscript{2}},
 \textbf{James Flemings\textsuperscript{2}},
\\
 \textbf{Krishna P. Gummadi\textsuperscript{1}},
 \textbf{Willie Neiswanger\textsuperscript{2}},
 \textbf{Robin Jia\textsuperscript{2}}
\\
 \textsuperscript{1}Max Planck Institute for Software Systems,
 \textsuperscript{2}University of Southern California
\\
 \small{
   \textbf{Correspondence:} \href{mailto:afkhan@mpi-sws.org}{afkhan@mpi-sws.org},
   \href{mailto:ameyagod@usc.edu}{ameyagod@usc.edu}
 }
}

\newcommand{\LibraryName}{TokenSmith }
\newcommand{\LibraryNameWithoutSpaceAtEnd}{TokenSmith}

\begin{document}
\maketitle
\begin{abstract}

% \james{Is it worth writing a sentence or two motivating \LibraryNameWithoutSpaceAtEnd? For example, it is important to understand the relationship between the training data and model behavior during pre-training; however, the current workflow makes it cumbersome and difficult to accomplish this.} 

Understanding the relationship between training data and model behavior during pretraining is crucial, but existing workflows make this process cumbersome, fragmented, and often inaccessible to researchers. We present \LibraryNameWithoutSpaceAtEnd, an open-source library for interactive editing, inspection, and analysis of datasets used in Megatron-style pretraining frameworks such as GPT-NeoX, Megatron, and NVIDIA NeMo. \LibraryName supports a wide range of operations including searching, viewing, ingesting, exporting, inspecting, and sampling data, all accessible through a simple user interface and a modular backend. It also enables structured editing of pretraining data without requiring changes to training code, simplifying dataset debugging, validation, and experimentation. \LibraryName is designed as a plug-and-play addition to existing large language model pretraining workflows, thereby democratizing access to production-grade dataset tooling. 

\LibraryName is hosted on GitHub\footnote{\url{https://github.com/aflah02/TokenSmith}}, with accompanying documentation and tutorials\footnote{\url{https://aflah02.github.io/TokenSmith/}}. A demonstration video is also available on YouTube.\footnote{\url{https://www.youtube.com/watch?v=cDO8VE9fZvU}}

\end{abstract}

\section{Introduction}

The barrier to pretraining large language models from scratch has been rapidly declining, driven by improved access to GPUs, a growing number of open-source frameworks, and the widespread sharing of technical knowledge. As a result, academic groups, open-source organizations and hobbyists are increasingly able to conduct meaningful pretraining research \cite{pythia, azerbayev2023proofnetautoformalizingformallyproving, chi-etal-2023-dissecting, forge, gupta2023continualpretraininglargelanguage, horawalavithana-etal-2022-foundation, ibrahim2024simplescalablestrategiescontinually, gao2025metadataconditioningaccelerateslanguage, gao2025trainlongcontextlanguagemodels, zeng2024turnwasteworthrectifying}.

However, a persistent challenge across existing frameworks is the lack of robust tooling for inspecting and interacting with the training data. 
% As the importance of "looking at your data" becomes clearer, we introduce \LibraryNameWithoutSpaceAtEnd, a toolkit designed to make this process seamless. Tasks such as debugging loss spikes by tracing relevant datapoints, generating modified datasets for counterfactual experiments or decontamination, and even viewing specific batches or sequences remain cumbersome in open-source setups \james{Could we briefly exemplify the current workflow for one of these tasks? I like the text about the current workflow for editing a dataset to produce a counterfactual version in the experiments section}. \LibraryName addresses these gaps by providing intuitive abstractions for editing, inspecting, and managing datasets thereby enabling faster iteration and deeper insight throughout the pretraining workflow.
Tasks such as debugging loss spikes by tracing relevant datapoints, generating modified datasets for counterfactual experiments or decontamination, and even viewing specific batches or sequences remain cumbersome in open-source setups. For example, producing a counterfactual dataset typically requires manually identifying files, ensuring token alignment, and re-tokenizing the entire corpus (a process that can take over a day for large datasets). 

We introduce \LibraryNameWithoutSpaceAtEnd, a toolkit designed to make this process seamless. 
\LibraryName addresses these gaps by providing intuitive abstractions for editing, inspecting, and managing datasets, thereby enabling faster iteration and deeper insight throughout the pretraining workflow.
% TokenSmith is built on top of Megatron-LM \cite{megatronLM}, a widely adopted and scalable framework for large language model pretraining. Notably, popular libraries such as GPT-NeoX \cite{gpt-neox-library} and NVIDIA NeMo\footnote{\url{https://github.com/NVIDIA/NeMo}} are also based on Megatron-LM and follow a common dataset format and training pipeline structure. There also exist several forks of Megatron which directly use it to train models such as K2\footnote{\url{https://github.com/LLM360/k2-train}} \citep{liu2025llm360k2building65b}, MAP-NEO\footnote{\url{https://github.com/multimodal-art-projection/Megatron-LM-NEO}} \citep{zhang2024mapneohighlycapabletransparent}, Megatron-Llama \footnote{\url{https://github.com/alibaba/Megatron-LLaMA}} and Apertus\footnote{\url{https://huggingface.co/collections/swiss-ai/apertus-llm-68b699e65415c231ace3b059}}. The framework is also used to share new efficient training methods \citep{qi2023zerobubblepipelineparallelism, qi2024pipelineparallelismcontrollablememory, ao2024seq1f1b, AcceleratingUSENIX, wan2025pipeoffloadimprovingscalabilitypipeline}.\footnote{\url{https://github.com/thunlp/Seq1F1B}}\footnote{\url{https://github.com/sail-sg/zero-bubble-pipeline-parallelism}}\footnote{\url{https://github.com/kwai/Megatron-Kwai}} The library is also adopted by other GPU providers such as AMD.\footnote{\url{https://github.com/ROCm/Megatron-LM}}

TokenSmith is built on top of Megatron-LM \cite{megatronLM}, a widely adopted and scalable framework for large language model pretraining. Several popular libraries such as GPT-NeoX \cite{gpt-neox-library} and NVIDIA NeMo\footnote{\url{https://github.com/NVIDIA/NeMo}} are also based on Megatron-LM, sharing its dataset format and training pipeline structure. Beyond these, a number of direct forks leverage Megatron-LM for model training, such as K2\footnote{\url{https://github.com/LLM360/k2-train}} \citep{liu2025llm360k2building65b}, MAP-NEO\footnote{\url{https://github.com/multimodal-art-projection/Megatron-LM-NEO}} \citep{zhang2024mapneohighlycapabletransparent}, Megatron-Llama\footnote{\url{https://github.com/alibaba/Megatron-LLaMA}}, and Apertus\footnote{\url{https://huggingface.co/collections/swiss-ai/apertus-llm-68b699e65415c231ace3b059}}. The framework also serves as a foundation for research into more efficient training methods \citep{qi2023zerobubblepipelineparallelism, qi2024pipelineparallelismcontrollablememory, ao2024seq1f1b, AcceleratingUSENIX, wan2025pipeoffloadimprovingscalabilitypipeline}, with public implementations available.\footnote{\url{https://github.com/thunlp/Seq1F1B}}$^{,}$\footnote{\url{https://github.com/sail-sg/zero-bubble-pipeline-parallelism}}$^{,}$\footnote{\url{https://github.com/kwai/Megatron-Kwai}} Finally, adoption extends beyond NVIDIA GPUs, as GPU providers such as AMD also maintain Megatron-LM support.\footnote{\url{https://github.com/ROCm/Megatron-LM}}

This shared foundation allows TokenSmith to natively support all three frameworks and other forks with minimal integration overhead. Additionally, TokenSmith is designed to be extensible, making it easy to add support for other frameworks.

\section{Library Offerings}

We build \LibraryName for practitioners and researchers working directly with large-scale language model pretraining. Our goal is to make it significantly easier to address a wide range of research questions and engineering challenges that arise when working with massive datasets and opaque training processes. Instead of offering a monolithic interface, we emphasize modularity and extensibility, providing intuitive abstractions through a simple frontend and a well-documented backend that can be adapted to different workflows. Here we describe the key functionalities we support: Inspect, Sample, Edit, Export, Ingest, and Search.

\subsection{Inspecting and Sampling Datasets}
\label{sec:inspecting-sampling}

Understanding the relationship between training data and model behavior is a recurring challenge in large-scale pretraining. Practitioners often need to trace issues back to specific sequences or isolate dataset subsets for hypothesis testing. Typical questions include:

\begin{itemize}
\item \textit{How can we trace and identify sequences that correlate with sudden spikes in training loss?}
\item \textit{Between two model checkpoints, what new data did the model encounter, and how might it explain improvements or regressions in performance?}
\item \textit{Are there tokenization issues or formatting inconsistencies that may have gone unnoticed?}
\item \textit{What happens if the model is trained only on a specific subset, such as domain-specific documents or repeated early-stage data?}
\item \textit{Can we sample sequences based on properties such as length, content patterns, or document metadata?}
\end{itemize}

\LibraryName provides a unified set of tools to support both deep inspection and flexible sampling:

\begin{itemize}
\item \textbf{Precise inspection utilities} allow users to locate and analyze data at the level of individual sequences, batches, or training steps using indices or global step numbers.
\item \textbf{Sampling utilities} support extracting subsets of the data based on custom policies, enabling rapid prototyping, ablation studies, and behaviorally targeted dataset construction.
\item \textbf{Modular integration} through a backend API allows seamless incorporation into training pipelines, analysis scripts, or interactive UIs.
\end{itemize}

By enabling structured, reproducible interrogation of the training dataset, \LibraryName empowers practitioners to move from anecdotal debugging to systematic, data-driven understanding of model behavior.

\subsection{Editing Datasets}
\label{sec:editing-utils}

As training progresses or evaluation findings emerge, practitioners frequently encounter the need to modify the dataset. These changes are often motivated by new insights or requirements, such as:

\begin{itemize}
\item \textit{Removing specific batches that correlate with spikes in training loss}
\item \textit{Filtering out examples that may result in test set leakage or data contamination}
\item \textit{Creating counterfactual variants of the dataset for controlled experiments, such as ablation studies or robustness analysis}
\end{itemize}

To support such use cases, \LibraryName offers a flexible editing interface that allows for \textbf{targeted edits}, enabling users to directly specify and modify individual sequences.

These capabilities allow researchers to iterate on dataset versions without re-engineering the training pipeline. Edits can be performed programmatically and are fully compatible with the inspection and sampling modules of the library. This makes it possible to run sophisticated data-centric experiments, such as testing the effect of subtle perturbations, while maintaining full control over what the model sees during training.

\subsection{Exporting Datasets}

Beyond sampling, reproducibility and interoperability are essential in dataset-centric research. Some recurring challenges are: 

\begin{itemize}
\item \textit{How can we verify and share a specific version of the dataset used for a paper, in a reproducible format compatible with popular libraries like HuggingFace Datasets?}
\item \textit{Can we export only specific batches/sequences of a dataset that show interesting trends to avoid sharing large binaries?}
\end{itemize}

To this end, \LibraryName includes \textbf{export tools} for converting datasets (entirely or in parts) into formats such as JSONL and CSV which are also HuggingFace compatible. These tools enable seamless sharing, integration with external pipelines, and long-term reproducibility of experimental results. 

\subsection{Ingesting Datasets}

Curating datasets for large-scale pretraining often begins with converting diverse data sources into a format compatible with Megatron-style frameworks. However, this step is frequently under-documented and error-prone. A common challenge is: \textit{How can we ingest and tokenize new datasets into the Megatron binary format without writing custom conversion pipelines?}

\LibraryName addresses this through streamlined \textbf{ingestion utilities} that support converting standard formats such as JSONL and CSV into the required .bin/.idx representation. These tools reduce the overhead of dataset preparation and ensure seamless compatibility with Megatron-based pretraining workflows.

\subsection{Searching Datasets}

As pretraining datasets grow in size and complexity, being able to efficiently search and retrieve relevant content becomes essential for both debugging and targeted experimentation. Practitioners and researchers often encounter challenges such as:

\begin{itemize}
\item \textit{How can we locate all occurrences of a specific phrase, token, or n-gram to inspect or remove sensitive or duplicate content?}
\item \textit{Can we curate the likely continuations given a naive n-gram model to contrast the generation likelihoods of our LLMs?}
\item \textit{Is it possible to trace model behaviors to specific textual patterns or domains within the training set?}
\item \textit{How do we efficiently support search at scale without loading the entire dataset into memory?}
\end{itemize}

To address these challenges, \LibraryName builds abstractions over Tokengram\footnote{\url{https://github.com/EleutherAI/tokengrams}}, an efficient n-gram indexing and search tool optimized for large-scale corpora. This allows users to perform fast searches over pre-tokenized corpora. Integrating Tokengram in \LibraryName provides support for end-to-end data interventions. For example, the search results can be processed with the Inspect and Export utilities for easy sharing with your collaborators or downstream post-processing. The matched documents from the search results can be processed with the Edit utilities. This might be useful to mask out toxic text, anonymize documents in place, etc.

By making dataset search fast and programmatically accessible, \LibraryName empowers users to build more informed training sets, track down model behaviors to specific training signals, and conduct controlled data-centric research at scale.

\section{Practical Case Studies}

In this section, we provide examples of how \LibraryName could simplify pipelines for NLP research.

\subsection{Training Dynamics of Memorization}

Several research groups have attempted to study memorization of natural~\cite{huang2024demystifyingverbatimmemorizationlarge,jagielski2023measuring} or counterfactually curated~\citep{fact-acquisition-kaist-10.5555/3737916.3739855,wei-etal-2024-proving} data during LM pre-training. In order to study training dynamics, these projects relied on one of two approaches:

\begin{enumerate}[leftmargin=\parindent]
\item \textbf{Re-tokenizing the corpus:} \citet{wei-etal-2024-proving} studied the memorization of watermarks in pre-training by injecting randomized strings in groups of documents. They re-tokenize the entire corpus along with different sets of watermarked documents. Note that the number of modified documents is a small fraction of the full corpus; thus, they spend considerable time re-tokenizing unchanged data. Moreover, their approach cannot control the order of unchanged documents in different training runs.
\item \textbf{Modifying the training library}: \citet{huang2024demystifyingverbatimmemorizationlarge} and \citet{fact-acquisition-kaist-10.5555/3737916.3739855} study verbatim memorization of documents in the early, middle, and late stages of pre-training by injecting curated/synthetically generated documents in specific training sequences. They achieve this by modifying the data loader and training loop of the pre-training library \citep{gpt-neox-library,olmo20242olmo2furious}. This engineering-intensive approach requires a deep understanding of the underlying pre-training libraries. Moreover, this may decrease the training efficiency of the library.
\end{enumerate}

In contrast, \LibraryName allows you to directly edit the tokenized dataset (\S~\ref{sec:editing-utils}). This allows you to perform the same experiments (1) without having to re-tokenize documents that haven't changed between training runs, and (2) without modifying the pre-training libraries.

\subsection{Identifying Causes of Instability}

\citet{olmo20242olmo2furious} highlight that sudden spikes in training loss can lead to instability further along in pre-training and worse final model performance. In order to debug the cause of the instability, they inspect the batches of data that caused the spikes and identify that the batches contain training sequences with repeated n-grams.
% Recreating this would require modifying the training library to dump the current batch of data if there is a spike in the loss. If the batch number where the loss spike occurred is known (from a previous training run), they may run the data loader up to the identified batch number and then dump the next batch of data.
The Inspect and Export tools in \LibraryName provide a straightforward interface to extract batches based on the step number (where the loss spike occurred).

\section{Library Design}

\LibraryName is designed to support two complementary modes of interaction: a Pythonic API for seamless integration into existing training or analysis pipelines, and a visual UI for interactive exploration and inspection. 

\subsection{Overview: Megatron Dataset Format}

% Megatron-LM (Megatron-Core) builds on a multi-layer dataset abstraction. At the lowest level is the IndexedDataset (and its builder) that holds token IDs in a binary .bin file and metadata in a .idx file. The .idx file first records dataset-level metadata (counts, offsets), then sequence-level info (lengths, byte offsets, document boundaries)\footnote{\url{https://docs.nvidia.com/megatron-core/developer-guide/latest/api-guide/datasets.html}}.

Megatron-LM’s indexed format uses two files per data split. The .bin file contains raw token sequences (packed back-to-back) as a flat array of integers. The .idx file contains metadata and pointers into the .bin. Specifically, the index begins with a header (version, dtype, number of sequences, number of documents) and then lists, for each sequence, its length (number of tokens) and its byte offset in the .bin. It also records, for each document, which sequences belong to it. In effect, .idx lets the dataset class reconstruct which slice of the big token array corresponds to each example.

The consistency of this format across libraries allows \LibraryName to support them out-of-the-box with minimal adaptations, enabling seamless interoperability and inspection without requiring separate data handling logic for each framework.

\subsection{Pythonic API}

\begin{figure}[h!]
    \centering
    \includegraphics[width=0.48\textwidth]{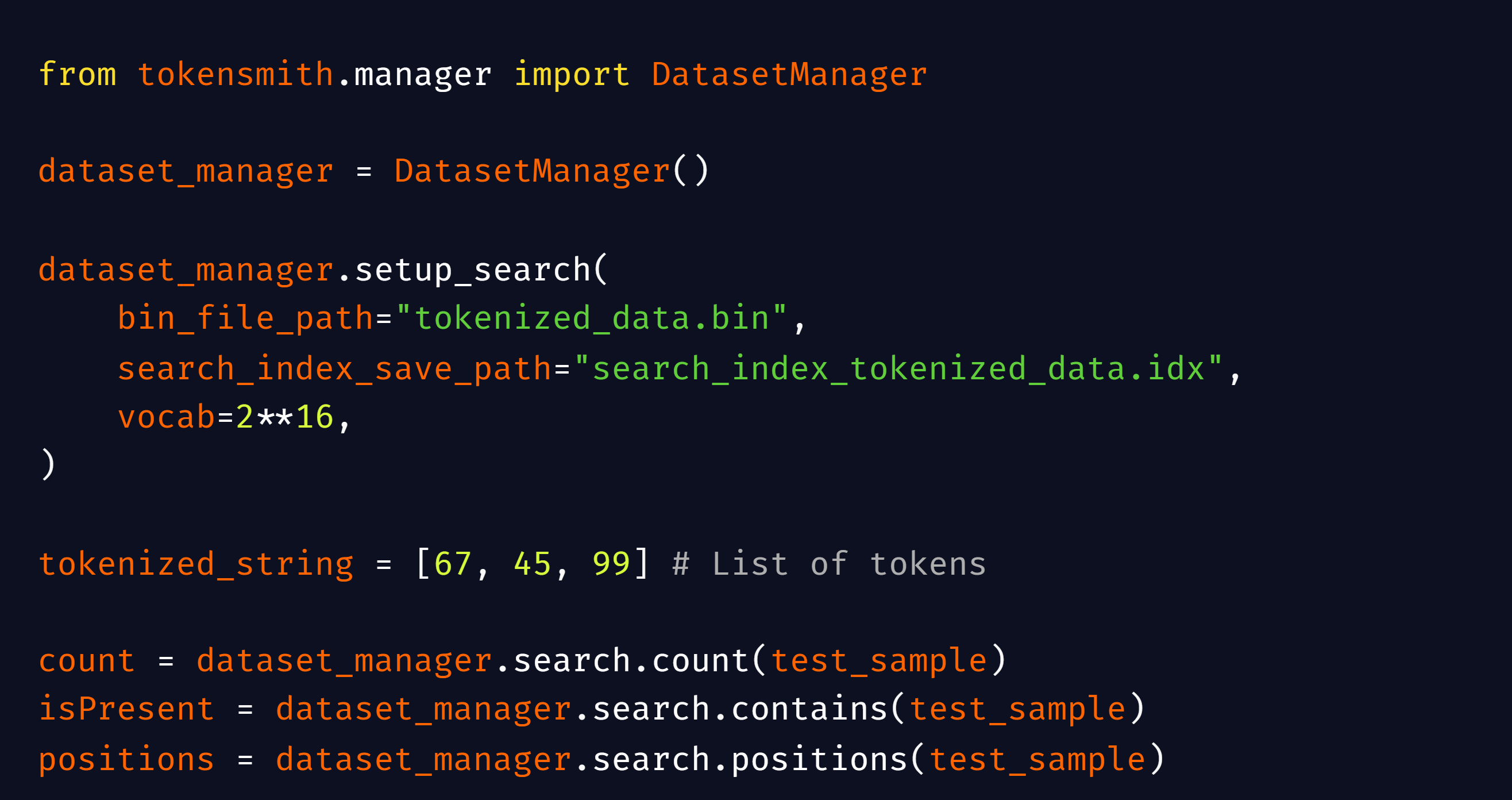}
    \caption{Search API}
    \label{fig:search_api}
\end{figure}

\begin{figure}[h!]
    \centering
    \includegraphics[width=0.48\textwidth]{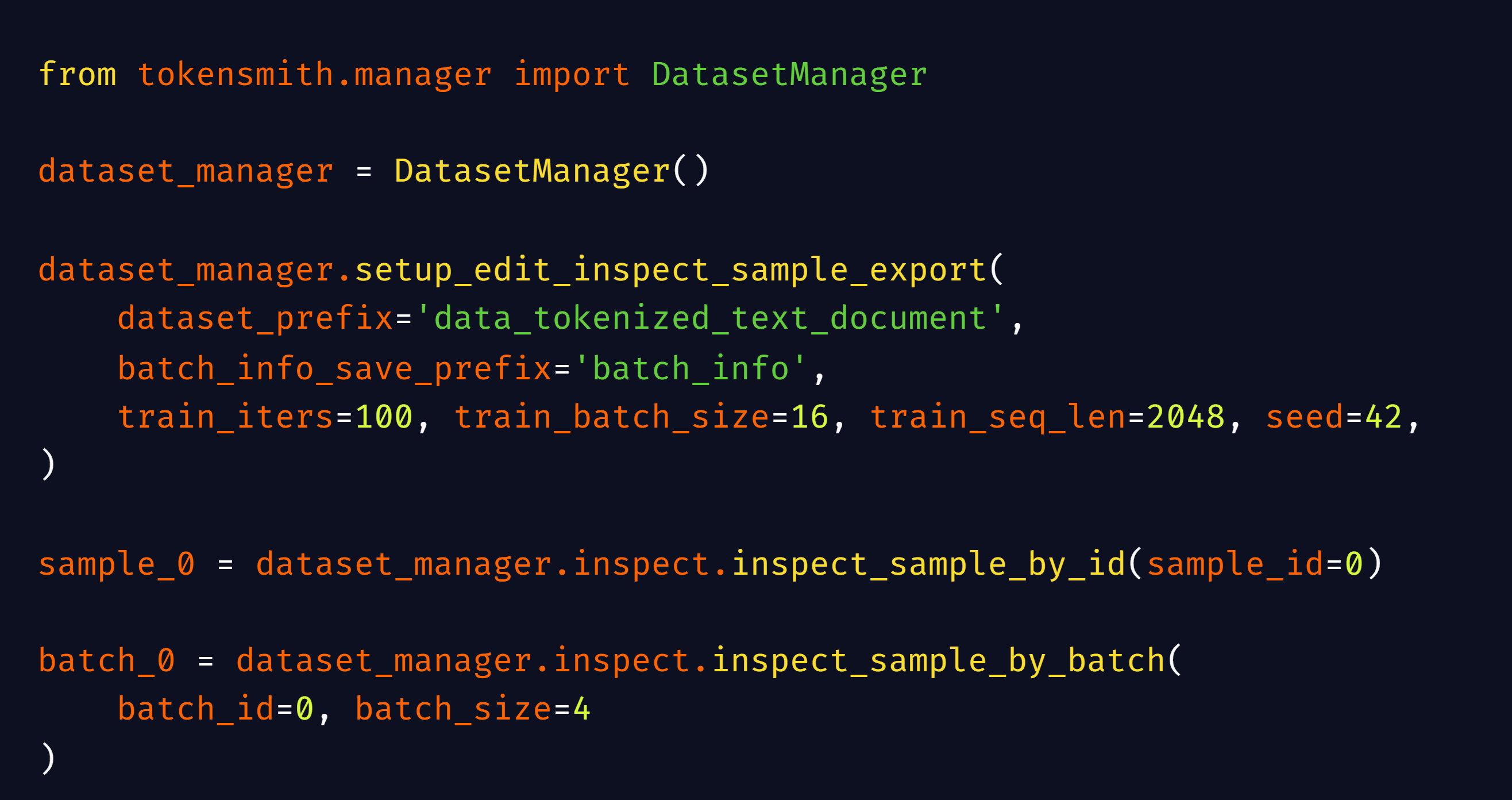}
    \caption{Inspect API}
    \label{fig:inspect_api}
\end{figure}

\begin{figure}[h!]
    \centering
    \includegraphics[width=0.48\textwidth]{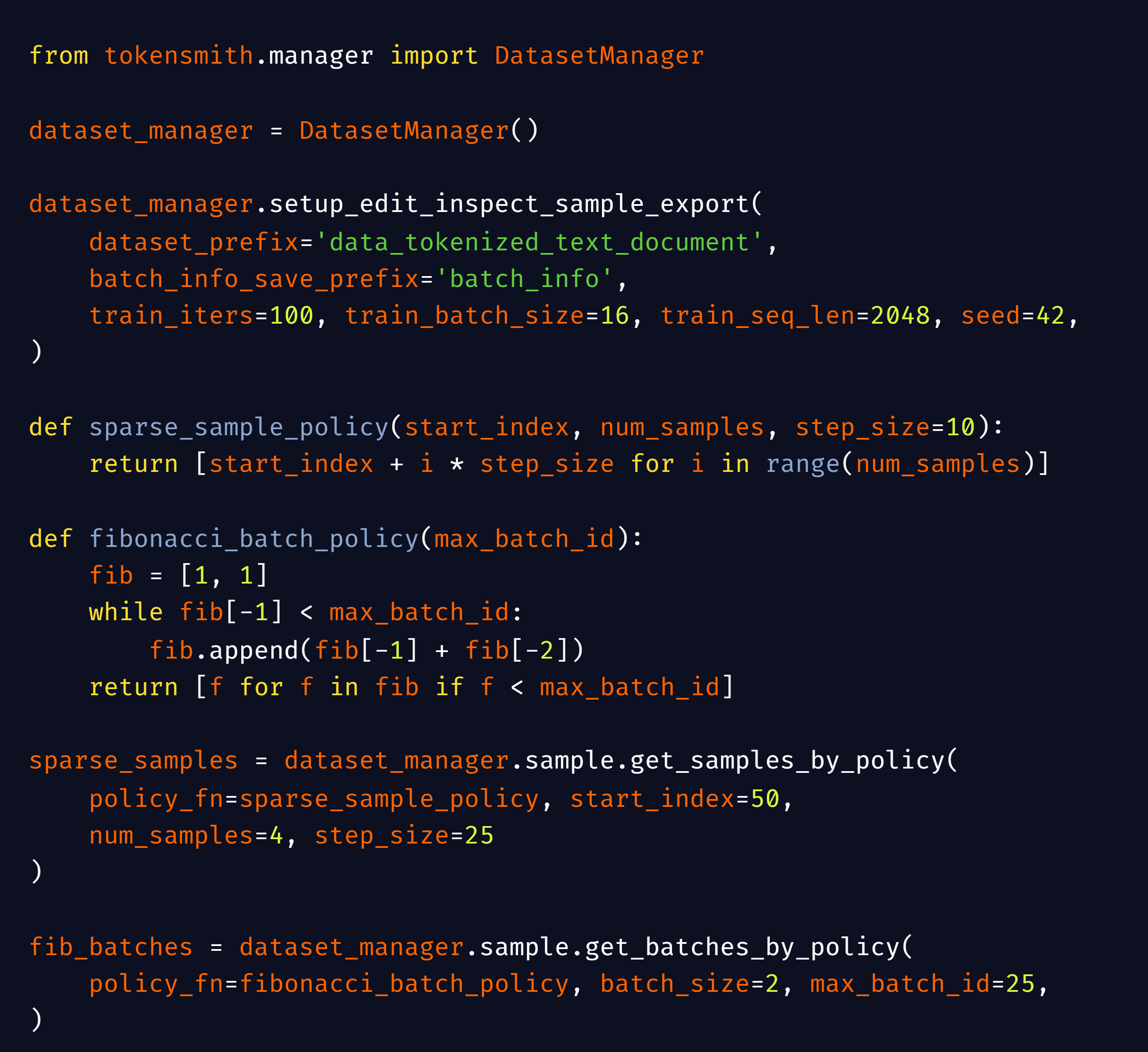}
    \caption{Sample API}
    \label{fig:sampling_api}
\end{figure}

\begin{figure}[h!]
    \centering
    \includegraphics[width=0.48\textwidth]{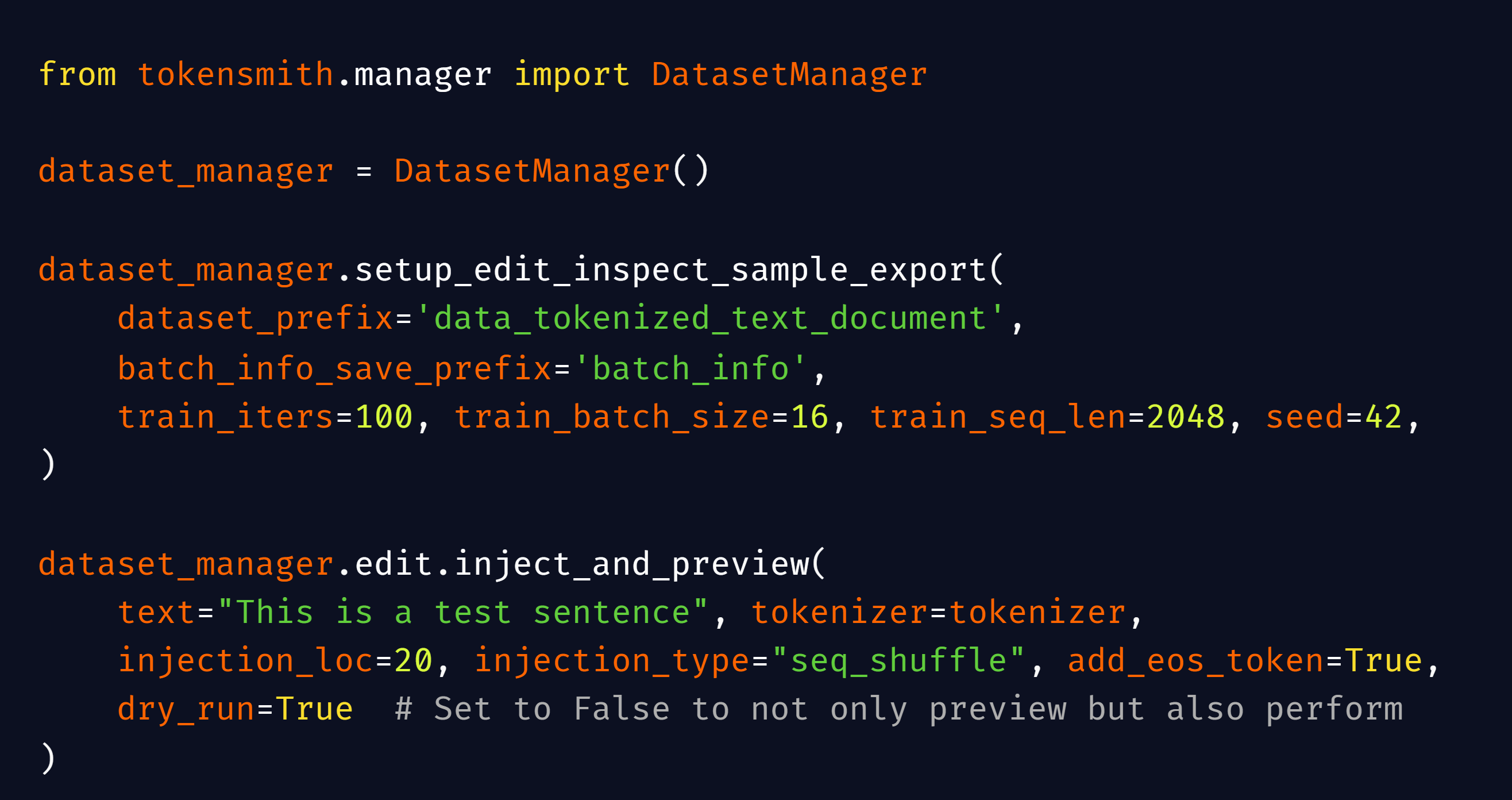}
    \caption{Edit API}
    \label{fig:edit_api}
\end{figure}

\begin{figure}[h!]
    \centering
    \includegraphics[width=0.48\textwidth]{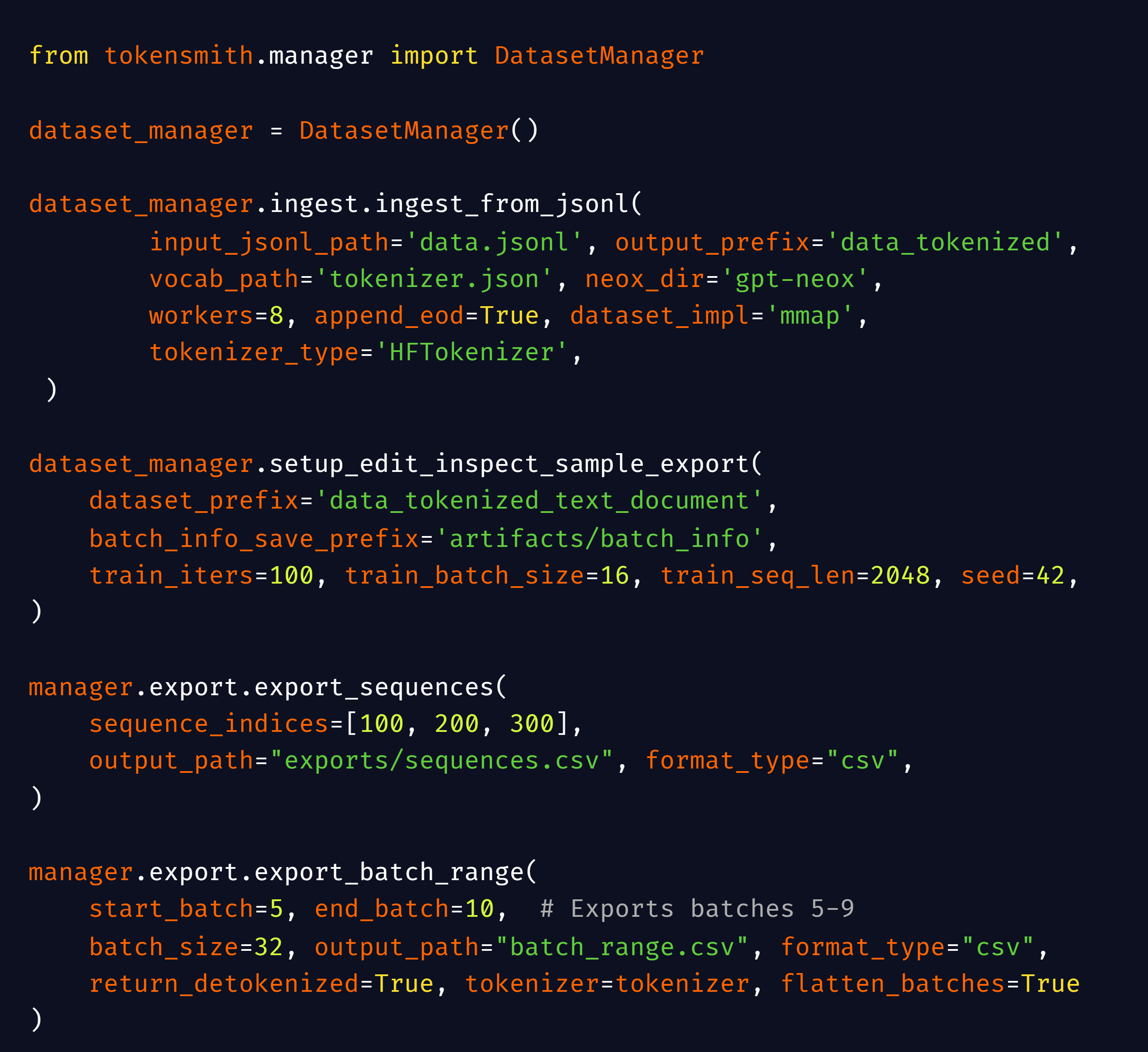}
    \caption{Ingest and Export API}
    \label{fig:ingest_export}
\end{figure}

\LibraryName exposes a modular, object-oriented API that allows users to programmatically ingest, edit, search, sample, inspect, and export datasets. This API is well-suited for integration into training scripts, research notebooks, or batch processing workflows. Figures~\ref{fig:search_api}, \ref{fig:inspect_api}, \ref{fig:sampling_api},  \ref{fig:edit_api} and \ref{fig:ingest_export} illustrate the core API patterns:

\begin{itemize}
\item \textbf{Figure~\ref{fig:search_api}} illustrates how users can configure and execute token-level search queries using a Tokengram-backed index over the dataset.
\item \textbf{Figures~\ref{fig:inspect_api} and \ref{fig:sampling_api}} showcase the interfaces for inspecting specific sequences and sampling data according to user-defined policies.
\item \textbf{Figure~\ref{fig:edit_api}} presents the editing interface, which supports fine-grained, targeted modifications to existing sequences.
\item \textbf{Figure~\ref{fig:ingest_export}} demonstrates the dataset ingestion and export utilities, which allow users to import new corpora and export subsets or modified datasets in standard formats.
\end{itemize}
The library is designed to latch onto your existing pretraining environments with minimal configuration (instructions outlined clearly in the README).\footnote{\url{https://github.com/aflah02/tokensmith?tab=readme-ov-file\#-quick-start}}

\subsection{Interactive UI}

\begin{figure}[h!]
    \centering
    \begin{subfigure}[t]{0.48\textwidth}
        \centering
        \includegraphics[width=\textwidth]{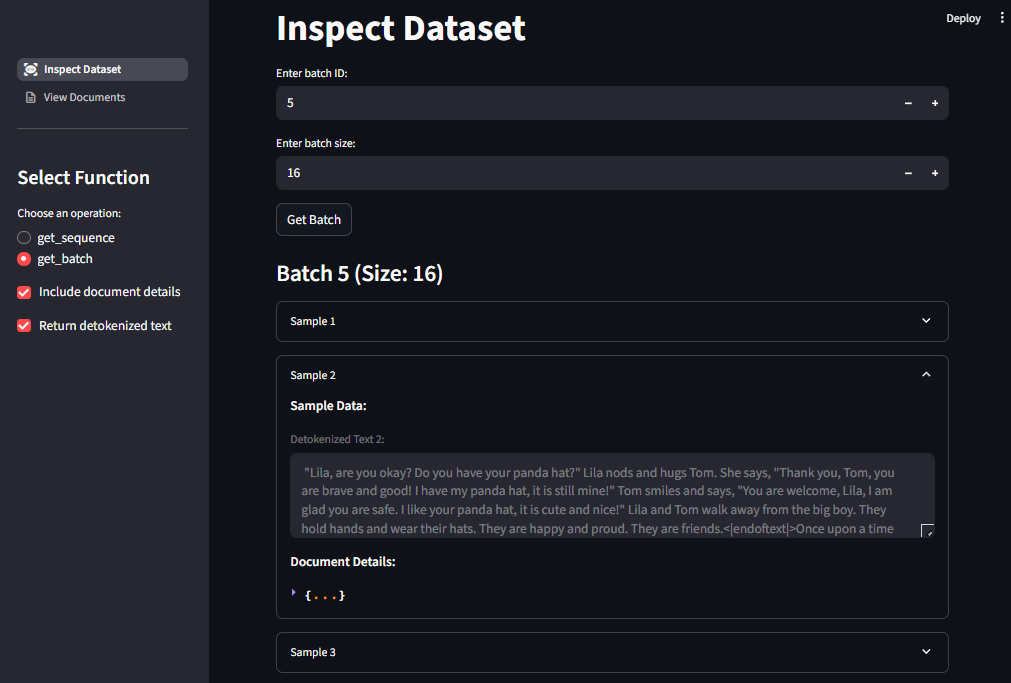}
        \caption{Inspect UI showing detailed information for a selected batch, including tokenized and detokenized sequences and document metadata.}
        \label{fig:inspect_ui}
    \end{subfigure}
    \hfill
    \begin{subfigure}[t]{0.48\textwidth}
        \centering
        \includegraphics[width=\textwidth]{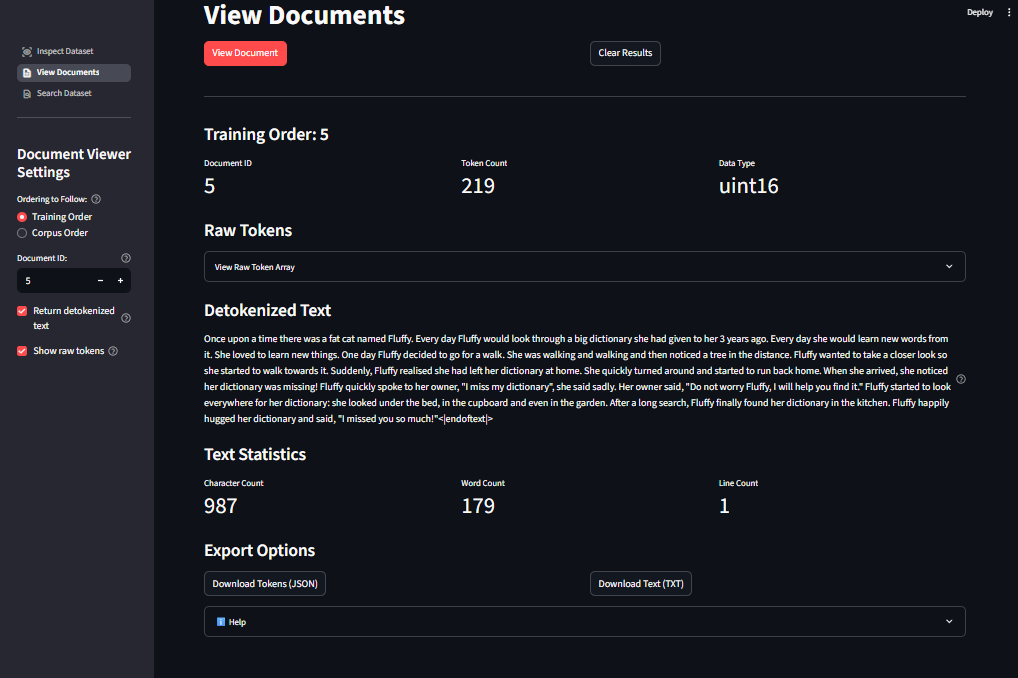}
        \caption{View Document UI for browsing individual documents and their tokenized representations.}
        \label{fig:view_doc_ui}
    \end{subfigure}
    \caption{TokenSmith inspection and viewing interfaces for exploring dataset contents at the batch and document levels.}
    \label{fig:inspect_view_ui}
\end{figure}

% To support visual exploration and rapid debugging, \LibraryName includes an intuitive user interface built with Streamlit.\footnote{\url{https://streamlit.io/}} This interface also serves as a reference implementation that can be extended or adapted to other frameworks or custom designs based on user requirements. This interface supports interactive search, batch-level inspection, and document editing capabilities. Figure~\ref{fig:inspect_ui} and \ref{fig:view_doc_ui} showcases one of the UI pages while Figures~\ref{fig:search_ui_grid} (in Appendix~\ref{app:ui-samples}) present example views of the search and view document tools, respectively. Users can browse through individual sequences and batches, for example, to trace those corresponding to steps with abnormal loss spikes. They can also query phrase occurrence counts, verify the presence of specific strings in the dataset, identify their exact locations, and retrieve the distribution of next token predictions, all through straightforward point and click interactions.

To enable visual exploration and rapid debugging, \LibraryName provides an intuitive user interface built with Streamlit.\footnote{\url{https://streamlit.io/}} The UI serves both as a reference implementation and a customizable layer for users to extend based on their workflows. It supports interactive search, batch and sequence level inspection, and document viewing. Figures~\ref{fig:inspect_ui} and \ref{fig:view_doc_ui} show examples of the inspect and document viewing pages, while Figure~\ref{fig:search_ui_grid} (Appendix~\ref{app:ui-samples}) illustrates different search modes. Users can browse individual sequences or batches to trace issues such as loss spikes, search for specific phrases, locate them within documents, and explore next token distributions, all through simple point and click interactions.

Together, the API and UI provide a unified and flexible interface to dataset management, enabling both hands-on experimentation and automated workflows at scale.

% A detailed overview of the design patterns used in building the library is outlined in Appendix~\ref{app:design_patterns}.

\subsection{Design Patterns}

\LibraryName is built with a strong emphasis on clean software engineering to ensure ease of use, extensibility, and long-term maintainability. Its architecture follows established design patterns to provide a clear separation of concerns, enable safe experimentation, and support both research and production environments. These design choices also make it easier for contributors to extend and integrate the toolkit with custom workflows. A detailed breakdown of the patterns used, including handler-based modularity, a facade interface, and runtime configurability, is provided in Appendix~\ref{app:design_patterns}.

\section{Benchmarking}

While there are no established baselines for many of the operations supported by \LibraryName we compare against the de facto workflows that practitioners currently rely on to achieve similar outcomes. These comparisons highlight the complexity and overhead of existing approaches, and demonstrate how \LibraryName streamlines them.

\begin{itemize}[leftmargin=\parindent]
\item \textbf{Editing a dataset to produce a counterfactual version:}
Consider the case where a researcher wants to generate a modified dataset that differs by only a few examples from an original corpus spanning hundreds of billions of tokens.
\begin{itemize}[leftmargin=\parindent]
\item \textit{Current workflow:} Manually identify and replace relevant files, ensure token alignment (if needed), and re-tokenize the entire dataset. This process is brittle and time-consuming; for example, tokenizing a 500B-token corpus can take over a day depending on the system configuration.
\item \textit{With \LibraryNameWithoutSpaceAtEnd:} Users can programmatically perform targeted or randomized edits directly on the tokenized dataset using our editor interface. This removes the need to reason about token boundaries or initiate a full re-tokenization pass, significantly reducing iteration time and engineering overhead.
\end{itemize}
\item \textbf{Sampling Sequences According to Custom Policies:}
\begin{itemize}[leftmargin=\parindent]
\item \textit{Without \LibraryName:} Practitioners must manually extract sequences from the binary files, align them with training indices, implement policy-based filtering logic, and reconstruct the final subset (often requiring non-trivial changes to the existing pipeline).
\item \textit{With \LibraryName:} The sampling API abstracts away these complexities, allowing users to specify high-level parameters and a custom policy function to obtain the desired subset with minimal effort.
\end{itemize}

\end{itemize}

\begin{figure}[h!]
    \centering
    \includegraphics[width=0.48\textwidth]{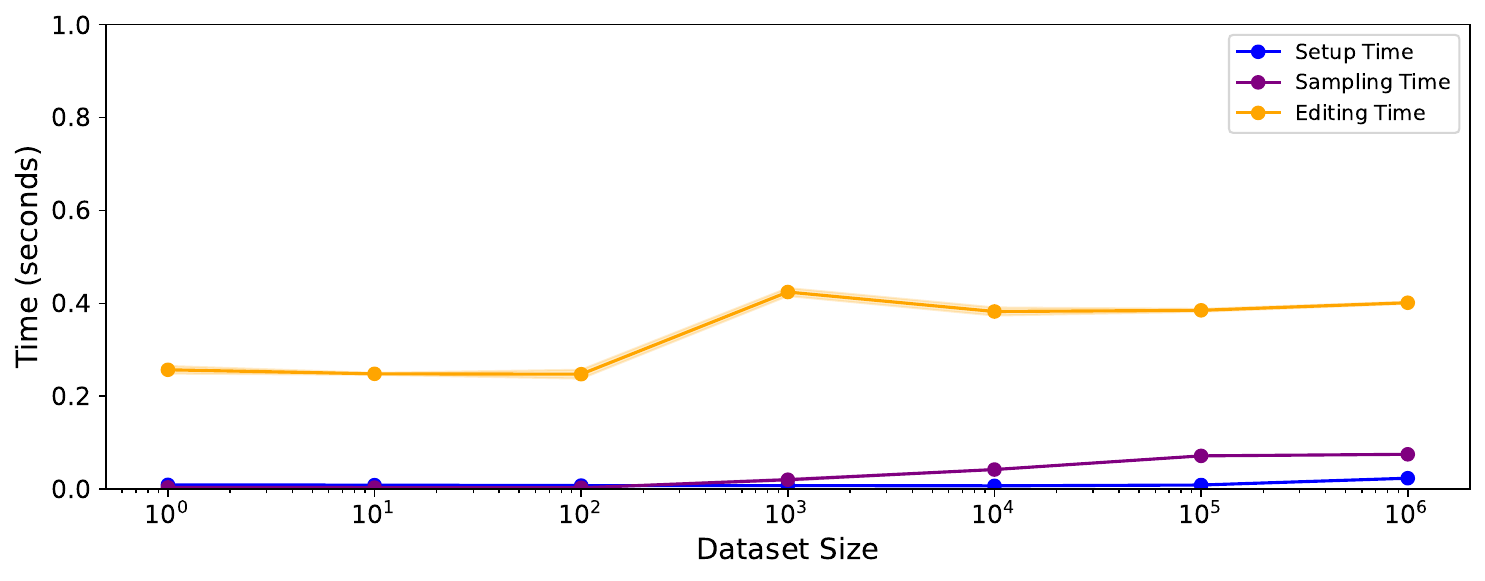}
    \caption{Execution time for setup, sampling, and editing operations across varying dataset sizes. For sampling and editing, the reported times correspond to 100 operations.}    \label{fig:combined}
\end{figure}

We also examine the scalability of our system by measuring the time taken for three representative operations as dataset size increases.

\subsection{Dataset Setup}

We measure the time required to execute the \texttt{setup\_edit\_inspect\_sample\_export} method, which initializes all necessary handlers based on the user’s configuration. This step introduces negligible overhead, remaining under 0.03 seconds even for corpora with one million documents.

\subsection{Sampling Performance}

We randomly sample 100 sequence indices and measure the total retrieval time. This process is repeated five times, and we report the average and standard deviation. A single-item fetch is performed beforehand to warm up the system. Sampling remains stable across dataset sizes. For example, sampling 100 sequences from a 1M-document corpus takes under 0.1 seconds.

\subsection{Editing Time}

To assess editing performance, we insert the sentence \texttt{(This is a test sentence.)} at 100 random positions across the dataset, following an initial warm-up edit. This operation is repeated five times, with average time and standard deviation reported. Edit latency remains under 0.5 seconds even for a 1M-document corpus, indicating minimal sensitivity to dataset size.

We showcase the benchmarking results in Figure~\ref{fig:combined}. These results highlight that \LibraryName offers low-latency interactivity, making it well-suited for both rapid experimentation and large-scale dataset manipulation. For search-related benchmarks, we refer readers to Tokengrams’ evaluations,\footnote{\url{https://github.com/EleutherAI/tokengrams?tab=readme-ov-file\#performance}} as TokenSmith directly integrates Tokengrams for all token-level search operations.

Before each benchmark measurement, we reinstantiate the \texttt{DatasetManager} and explicitly trigger garbage collection using \texttt{gc.collect()} from Python’s \texttt{gc} module.\footnote{\url{https://docs.python.org/3/library/gc.html}} The benchmarking script, along with the corresponding results, is available in the repository.\footnote{\url{https://github.com/aflah02/TokenSmith/tree/main/benchmarking}}

\section{Conclusion}

We present \LibraryNameWithoutSpaceAtEnd, a modular and extensible toolkit designed to streamline dataset-centric workflows in Megatron-style LLM pretraining. By offering intuitive abstractions for ingesting, editing, inspecting, sampling, searching, and exporting training data, TokenSmith fills a crucial gap in current open-source infrastructure. TokenSmith's support for multiple backends, efficient operations at scale, and dual interface (Pythonic API and visual UI) makes it accessible to researchers, practitioners, and hobbyists alike. As the LLM ecosystem increasingly embraces open and reproducible research, we believe \LibraryName will serve as a practical foundation for understanding, debugging, and experimenting with the data that drives modern language models.

% \textbf{Features:}

% \begin{itemize}
%     \item Fine-grained control over inspection, including the ability to view specific batches or isolate individual sequences based on sequence index, batch index, or global step.
%     \item A clean, minimal frontend for visualizing data, with a modular backend API that allows users to build custom interfaces or integrate with other tools.
% \end{itemize}

\section*{Acknowledgments}

We thank the EleutherAI team for open-sourcing Tokengrams and GPT-NeoX, and for their helpful responses to our questions. We also acknowledge the contributions of NVIDIA’s Megatron and GPT-NeoX repositories, which serve as foundational components in our work. 
The results presented in this work used compute resources from the National AI Research Resource Pilot, with support from NVIDIA, including NVIDIA's DGX Cloud product and the NVIDIA AI Enterprise Software Platform. This work was supported in part by a gift from the USC-Amazon Center on Secure and Trusted Machine Learning, and the National Science Foundation under Grant No. IIS-2403436. Any opinions, findings, and conclusions or recommendations expressed in this material are those of the author(s) and do not necessarily reflect the views of the National Science Foundation. 
Finally, we note that large language models were used to assist in editing and refining the writing of this paper.

\section*{Licensing}

\LibraryName is released under the Apache 2.0 license. This permissive license allows for both academic and commercial use, as well as modification and redistribution, making it suitable for a wide range of research and production workflows.

% Bibliography entries for the entire Anthology, followed by custom entries
%\bibliography{anthology,custom}
% Custom bibliography entries only
\bibliography{custom}

\appendix

\section{Design Patterns Employed in the Library}
\label{app:design_patterns}

\LibraryName is structured around well-established software design principles that promote modularity, extensibility, and maintainability. The internal architecture is intentionally clean and componentized to accommodate both research prototyping and production-scale workflows. Below, we describe the primary design patterns used in the codebase.

\paragraph{Handler Pattern (Command or Service Object) \cite{gamma1995design}}
Each major functional area (editing, inspecting, sampling, exporting, and searching) is encapsulated within its own handler class. For instance, EditHandler, InspectHandler, SampleHandler, ExportHandler, and SearchHandler each manage their domain-specific logic while exposing a consistent interface. This clear separation of concerns makes the system easier to extend, test, and reason about.

\paragraph{Facade Pattern \cite{gamma1995design}}
The DatasetManager class serves as a unified entry point to the system. It orchestrates the initialization of handlers and provides a high-level API to the end user. This shields users from internal complexities and reduces the cognitive load involved in accessing multiple capabilities.

\paragraph{Dependency Injection}
Rather than relying on tight coupling or global state, handlers receive references to the DatasetManager or its specific components during initialization. This inversion of control enhances testability and supports future decoupling and modular reuse.

\paragraph{Type Hinting and Forward References}
To avoid circular dependencies while retaining strong type safety, the library uses TYPE\_CHECKING blocks and string-based type annotations (e.g., 'DatasetManager'). This allows static analyzers and IDEs to provide full support while maintaining clear dependency boundaries.\footnote{\url{https://peps.python.org/pep-0484/}}

\paragraph{Strategy Pattern (Configurable Behavior) \cite{gamma1995design}}
Handlers expose methods whose behavior can be configured at runtime via parameters. For example, the EditHandler supports multiple injection strategies. This makes the system adaptable for experimentation without requiring internal changes to core logic.

\paragraph{Template Method Pattern \cite{gamma1995design}}
Export operations follow a template structure in which base methods (such as export) provide a standard workflow but allow subclasses or extensions to override certain steps. This approach encourages consistent behavior while allowing flexibility for future extensions or format support.

\paragraph{Validation and Defensive Programming}
Robust input validation and error handling are systematically applied across the codebase. Although not a formal design pattern, this practice contributes significantly to the reliability and maintainability of the library, especially in high-scale or adversarial settings.

\paragraph{Modular Package Structure}
The code is divided into clearly defined submodules (edit, inspect, sample, export, search), with each exposing a single handler class through its \_\_init\_\_.py. This supports the Single Responsibility Principle \cite{martin2003agile} and allows contributors to quickly locate, understand, and extend functionality.

These design decisions collectively ensure that \LibraryName remains extensible and maintainable as it grows to support additional backends, workflows, and research use cases.

\section{User Interface}
\label{app:ui-samples}

\begin{figure*}[h!]
    \centering
    \begin{subfigure}[t]{0.48\textwidth}
        \centering
        \includegraphics[width=\textwidth]{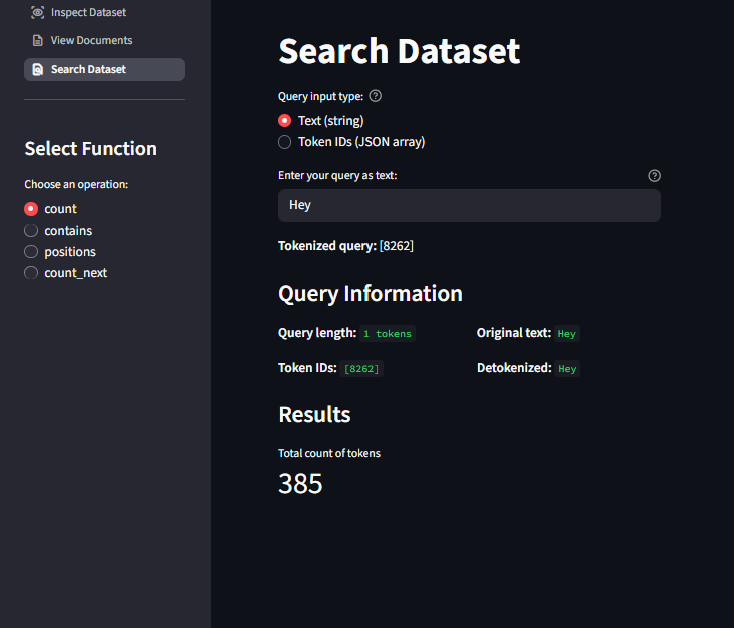}
        \caption{Search for count}
        \label{fig:search_count}
    \end{subfigure}
    \hfill
    \begin{subfigure}[t]{0.48\textwidth}
        \centering
        \includegraphics[width=\textwidth]{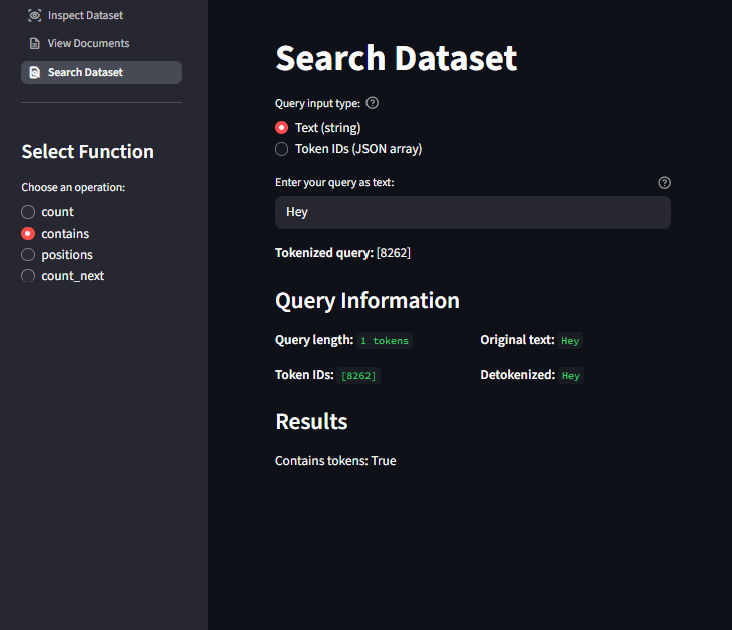}
        \caption{Search for presence}
        \label{fig:search_contains}
    \end{subfigure}

    \vspace{0.5em}

    \begin{subfigure}[t]{0.48\textwidth}
        \centering
        \includegraphics[width=\textwidth]{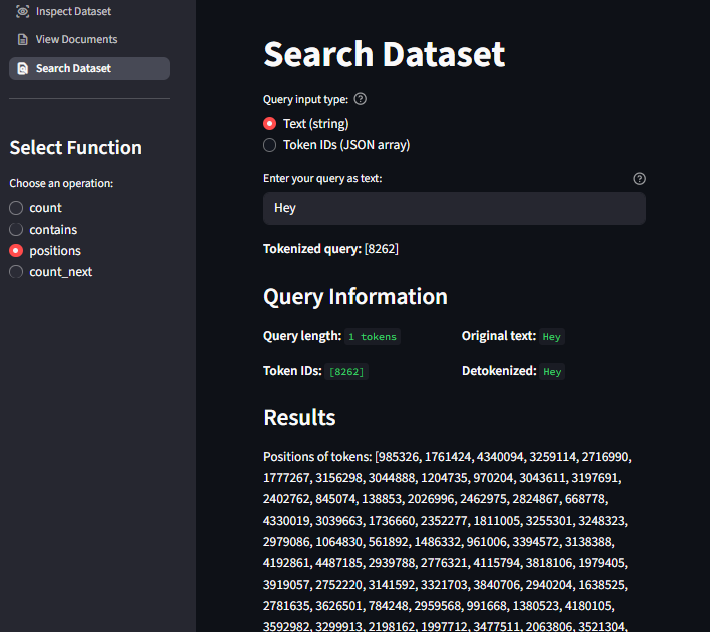}
        \caption{Search for positions of occurrence}
        \label{fig:search_position}
    \end{subfigure}
    \hfill
    \begin{subfigure}[t]{0.48\textwidth}
        \centering
        \includegraphics[width=\textwidth]{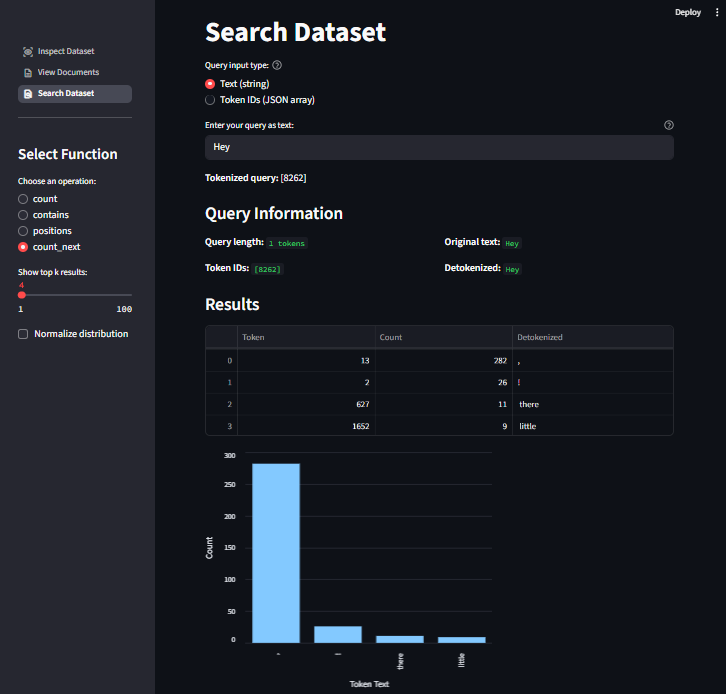}
        \caption{Search for likely next tokens using an n-gram model}
        \label{fig:search_next_token}
    \end{subfigure}

    \caption{TokenSmith search UI showing different search strategies: filtering by count, containment, positional match, and context-based continuation.}
    \label{fig:search_ui_grid}
\end{figure*}

Figure~\ref{fig:search_ui_grid} showcases different components of the \LibraryName search interface, illustrating how users can query token counts, presence, positions of occurrence, and likely next tokens using an n-gram model—all through intuitive visual tools.

\end{document}